\documentclass[10pt,twocolumn,letterpaper]{article}

\usepackage{cvm}
\usepackage{times}
\usepackage{epsfig}
\usepackage{graphicx}
\usepackage{amsmath}
\usepackage{amssymb}
\usepackage[normalem]{ulem}
\useunder{\uline}{\ul}{}
\usepackage{graphicx}

\usepackage[pagebackref=true,breaklinks=true,letterpaper=true,colorlinks,bookmarks=false]{hyperref}

\newcommand{\keywords}[1]{{\bf \emph{Keywords: #1}}}

\cvmfinalcopy

\def\cvmPaperID{****} 

\ifcvmfinal\fi
\begin{document}

\title{Data relativistic uncertainty framework for low-illumination anime scenery image enhancement}

\author{Yiquan Gao\thanks{First and corresponding author; independently led the project and manuscript preparation; funded solely by the author.}\\
 Heriot-Watt University\\
\and
 John See\thanks{Provided general feedback and editorial suggestions only.}\\
Heriot-Watt University\\
}

\maketitle

\begin{abstract}
	By contrast with the prevailing works of low-light enhancement in natural images and videos, this study copes with the low-illumination challenges in anime scenery images to bridge the domain gap. For such an underexplored enhancement task, we first curate images from various sources and construct an unpaired anime scenery dataset with diverse environments and illumination conditions to address the data scarcity. To exploit the power of uncertainty information inherent with the diverse illumination conditions, we propose a \textbf{D}ata \textbf{R}elativistic \textbf{U}ncertainty (DRU) framework, motivated by the idea from Relativistic GAN. By analogy with the wave-particle duality of light, our framework interpretably defines and quantifies the illumination uncertainty of dark/bright samples, which is leveraged to dynamically adjust the objective functions to recalibrate the model learning under data uncertainty. Extensive experiments demonstrate the effectiveness of DRU framework by training several versions of EnlightenGANs, yielding superior perceptual and aesthetic qualities beyond the state-of-the-art methods that are incapable of learning from data uncertainty perspective. We hope our framework can expose a novel paradigm of data-centric learning for potential visual and language domains. Code is available\footnote{\url{https://github.com/StudioYG/DRU_framework}}.
\end{abstract}


\begin{figure}[h]
	\centering
	\includegraphics[width=\linewidth]{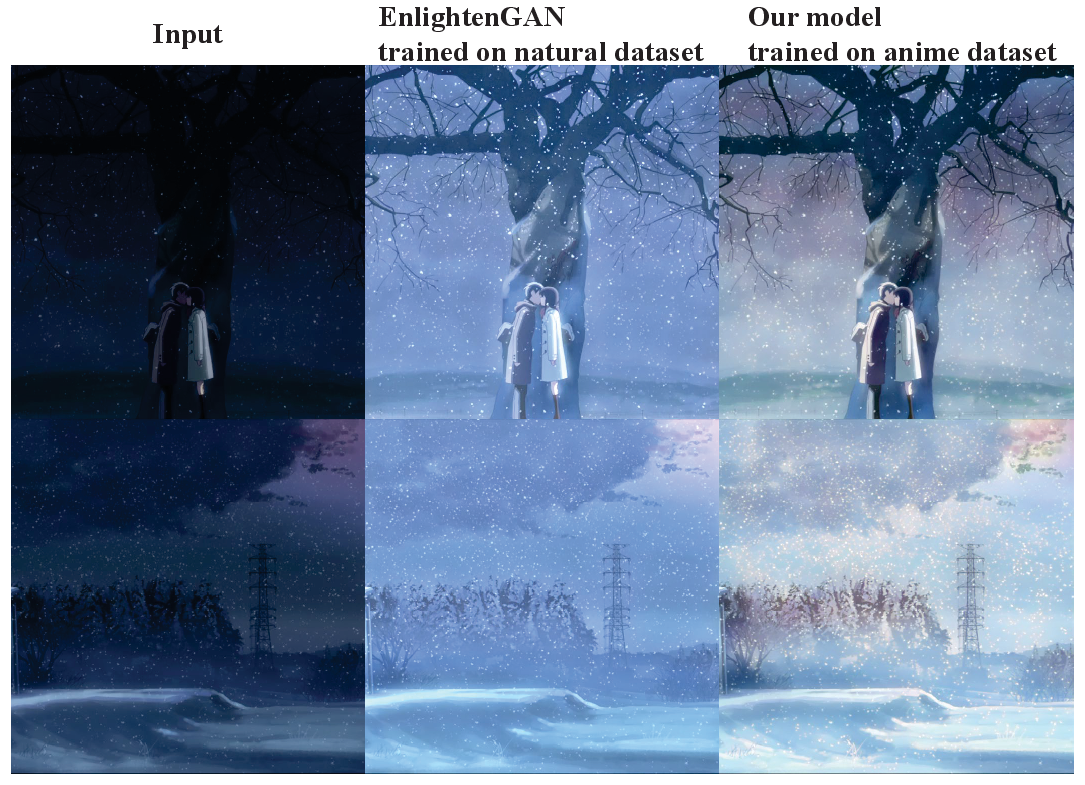}
	\caption{Domain gap between natural and anime image enhancement models: EnlightenGAN trained on natural data is prone to color bias and artifacts, such as the blue masks on enhanced results, while our model restores more aesthetically pleasing anime contents.}
	\label{Fig0}
\end{figure}

\section{Introduction}
In recent years, academic and industrial communities have raised a large number of attentions to the low-light enhancement (LLE) of natural images and videos~\cite{li2021low,rasheed2023comprehensive}. In contrast, the LLE task on anime scenery images has not been explored to date. Due to the artistic factor, the LLE task focuses on the perceptual and aesthetic qualities for anime scenery images, unlike mainly considering the perceptual quality for natural images. Anime scenery images are widely shared on the Internet and extensively used by general users for visual content creation. However, many such images suffer from reduced visibility and obscured fine details under low-illumination conditions, thereby limiting their direct usability. 

Although animation studios can enhance image quality by re-rendering the original artworks, this process requires substantial manual effort, and the original high-quality materials are generally inaccessible to the public. Moreover, low-illumination not only degrades perceptual quality but also negatively affects the robustness of downstream vision tasks on anime scenery images, including object detection, character segmentation, and diffusion-based image editing.

Therefore, developing effective low-illumination enhancement methods tailored to anime scenery images remains a practically important yet underexplored problem with broad real-world applicability.

Current natural image low-light enhancement methods exhibit undesirable color bias and artifacts (Fig.~\ref{Fig0}) when applied to anime scenery images, due to the domain gap problem~\cite{golyadkin2025closing,zhu2021mind}. Anime scenery images present distinct visual characteristics from natural images, making them effectively out-of-distribution data~\cite{farquhar2022out} for existing enhancement models.

Notably, no dedicated anime dataset exists for the low-illumination enhancement task. To this end, we collect anime scenery images from multiple open sources and develop a method to construct an unpaired anime scenery dataset using pseudo-data generation and a series of data processing techniques. This flexible data collection procedure extends the diversity of illumination conditions in the unpaired dark-bright images. 

A certain amount of illumination data is around the mid-brightness range with large variations, making it uncertain whether the image is dark or bright. In order to quantify the illumination uncertainty and enable the model to learn with the illumination uncertainty hidden in dark/bright images, we propose a data relativistic uncertainty (DRU) framework, which is inspired by the relativistic probability concept of RGANs~\cite{jolicoeur2018relativistic}. 

In specific, the framework defines the data uncertainty of dark/bright image illumination as the relativistic probability (RP) between an uncertain image and the ideal dark or bright image. We consistently treat each image of the unpaired anime scenery dataset as uncertain image including those images with confident illumination to predict their RPs, which are then adopted to dynamically adjust the training loss for recalibrating optimization with the uncertainty information from dark and bright images. 

Our work focuses on data-centric learning 
rather than on model-centric approaches that typically prioritize network architecture while largely neglecting data-level uncertainty~\cite{jiang2021enlightengan,liang2023iterative,li2021learning,liu2021retinex,ma2022toward}. We choose EnlightenGAN~\cite{jiang2021enlightengan} as the base architecture due to its significant advantages: I) multi-domain adaptability proven in real-world applications~\cite{mohamed2025integrating,wang2025unsupervised}; II) simple training scalability suitable for expanding data scale; III) lightweight unpaired training method friendly for running on most modern GPUs including 1080Ti and above. However, similar to most low-light enhancement methods, EnlightenGAN cannot exploit the data uncertainty inherent in dark or bright images, motivating our DRU framework.

Our contributions are highlighted as follows:
\begin{itemize}
	\item Different from the mainstream works on natural images and videos, we are the first to specifically focus on the low-illumination enhancement (LIE) for anime scenery images. Due to the lack of dedicated datasets for this task, we develop a data construction method by which the first unpaired anime scenery dataset is created, comprising of abundant dark/bright images with diverse scenes and illumination conditions to advance new studies in anime-related enhancement/synthesis.   
	\item We propose a novel data relativistic uncertainty (DRU) framework that introduces data uncertainty learning of dark/bright images into LIE. Taking advantage of the uncertainty information within the various illumination conditions, our framework intuitively enables the model training to allocate stronger optimization on illumination-confident samples and less on illumination-uncertain samples, leading to more adaptive performance on anime scenery images in different low illumination situations. 
	\item Extensive experiments show that our DRU-trained EnlightenGANs outperform state-of-the-art counterparts, proving the superiority of DRU. Ablation studies testify the importance of confident and uncertain images, and demonstrate the ability of DRU framework to alleviate the harm of training data noise caused by misclassification. We hope the data-centric paradigm of DRU can motivate new ideas of uncertainty learning in other vision or language tasks. 
\end{itemize}

\section{Related Works}

\subsection{Low-light Datasets} 
Natural low/normal-light image pairs are gathered by setting up one or several cameras in fixed positions and then adjusting the exposure time and ISO. This laborious collection setup constrains the diversity of data scenes and illumination conditions, making it hard to scale up the training pairs for supervised low-light image enhancement (LLIE). The commonly used paired datasets involve the first paired real-scene dataset LOL~\cite{Chen2018Retinex}, the multi-exposure datasets, e.g., SCIE~\cite{cai2018learning} and SID~\cite{Chen_2018_CVPR}, the photographic tone adjustment dataset broadly used in LLIE, i.e., MIT-Adobe FiveK~\cite{bychkovsky2011learning}. 

Unpaired datasets mainly consist of LIME~\cite{guo2016lime}, NPE~\cite{wang2013naturalness}, MEF~\cite{ma2015perceptual}, DICM~\cite{lee2013contrast} and VV~\cite{noauthor_vasileios_nodate}. Some unpaired datasets are collected for particular application scenarios, such as ExDARK~\cite{loh2019getting} for object detection, BBD-100K~\cite{yu2020bdd100k} for autonomous driving, DARK FACE~\cite{yang2020advancing} for face recognition. Due to the tedious collection setup, these datasets suffer from a few of drawbacks including limited data scale, duplicated scene and similar illumination variation, which can lead to certain learned bias or overfitting. Several studies~\cite{wang2023teaching,xu2022transformer,matsuura2023anicropify} show that anime images have different image characteristics from natural images. However, neither paired nor unpaired datasets exist to facilitate the low-illumination anime image enhancement task. Therefore, we build the first unpaired anime scenery dataset through curating multiple data sources from other studies~\cite{jiang2023scenimefy,chen2020animegan,zhu2017unpaired,skorokhodov2021aligning}. Note that compared to natural datasets, our anime dataset exhibits more diverse scene and illumination conditions across images since it is collected in a more flexible manner than the natural datasets.

\subsection{Unsupervised and Zero-Reference Enhancement}
Pairwise supervision easily causes the supervised learning~\cite{lore2017llnet,cai2023retinexformer,jiang2023low} to overfit the limited training sets as reported in ~\cite{wang2024zero}. To bypass the difficult-to-scalable pairwise training, unsupervised methods enable to train with merely unpaired low/normal-light images. EnlightenGAN~\cite{jiang2021enlightengan} is the first unsupervised learning method for LLIE, which employs global-local adversarial learning, self feature preserving loss and self-regularized attention mechanism to jointly guide its model training. NeRCo~\cite{yang2023implicit} proposes a dual-closed-loop cooperative constraint strategy to train the enhancement module concurrently with controllable fitting and multi-modal supervision components in a self-supervised manner. CLIP-LIT~\cite{liang2023iterative} introduces an iterative prompt learning strategy to alternatively optimize the prompt learning framework for finer prompts and train an unsupervised enhancement network using the improved text-image similarity. 

Zero-reference methods represent a higher-freedom form of unsupervised learning that does not require any paired or even unpaired data. Zero-DCE~\cite{guo2020zero} devises a series of non-reference loss functions to learn a lightweight enhancement network to estimate the parameters of a image-specific curve function, which is later refined by Zero-DCE++~\cite{li2021learning} for better inference speed and curve estimation. Motivated by the Retinex theory, a neural architecture search strategy called RUAS~\cite{liu2021retinex} explores LLIE network from a tiny search space, while requiring an iterative process for network searching. 

SCI~\cite{ma2022toward} establishes a self calibrated module to converge between results of each stage, enduing stronger representation into an individual basic block. A prior-to-image framework is proposed in~\cite{wang2024zero} to train a SD enhancement network using merely normal-light data, but the  physical quadruple prior is only applicable to natural images for the Kubelka–Munk theory and the model training is computationally intensive. 

Unsupervised and zero-reference methods often outperform the supervised counterparts in terms of the superior flexibility and generalization towards real-world applications~\cite{hu2021two,kim2022low,wang2025unsupervised}. Nevertheless, Zero-reference tasks may have difficulty in capturing reliable illumination concepts as they are learned without using any unpaired reference images. In addition, current research works are mainly model-centric. None of them attempts to explore the uncertainty information from the data with diverse illumination conditions for further data-centric learning. 

\subsection{Illumination Data Uncertainty}
Although illumination data itself carries a certain amount of uncertainty when it is classified as low or normal light, the potential of data uncertainty remains underexplored for LLIE task. Existing related works focus on bringing uncertainty into the model weights for adaptively tackling complex low-light data. Patch-wise-based diffusion model~\cite{li2025patch} applies the uncertainty weights from each patch as a guidance for the reverse diffusion process, therefore, balancing the abrupt regions of enhanced images. Retinex-based weight uncertainty network~\cite{jin2024weight} employs Bayesian neural network to incorporate uncertainty into the enhancement model by expressing each weight as a probability distribution instead of a scalar, thereby avoiding the overconfident enhancement from fixed weights. It can be seen that their research emphasis falls on the model uncertainty for LLIE, our work rather focuses on the potential of data uncertainty for LLIE. 

\begin{figure}[h]
	\centering
	\includegraphics[width=\linewidth]{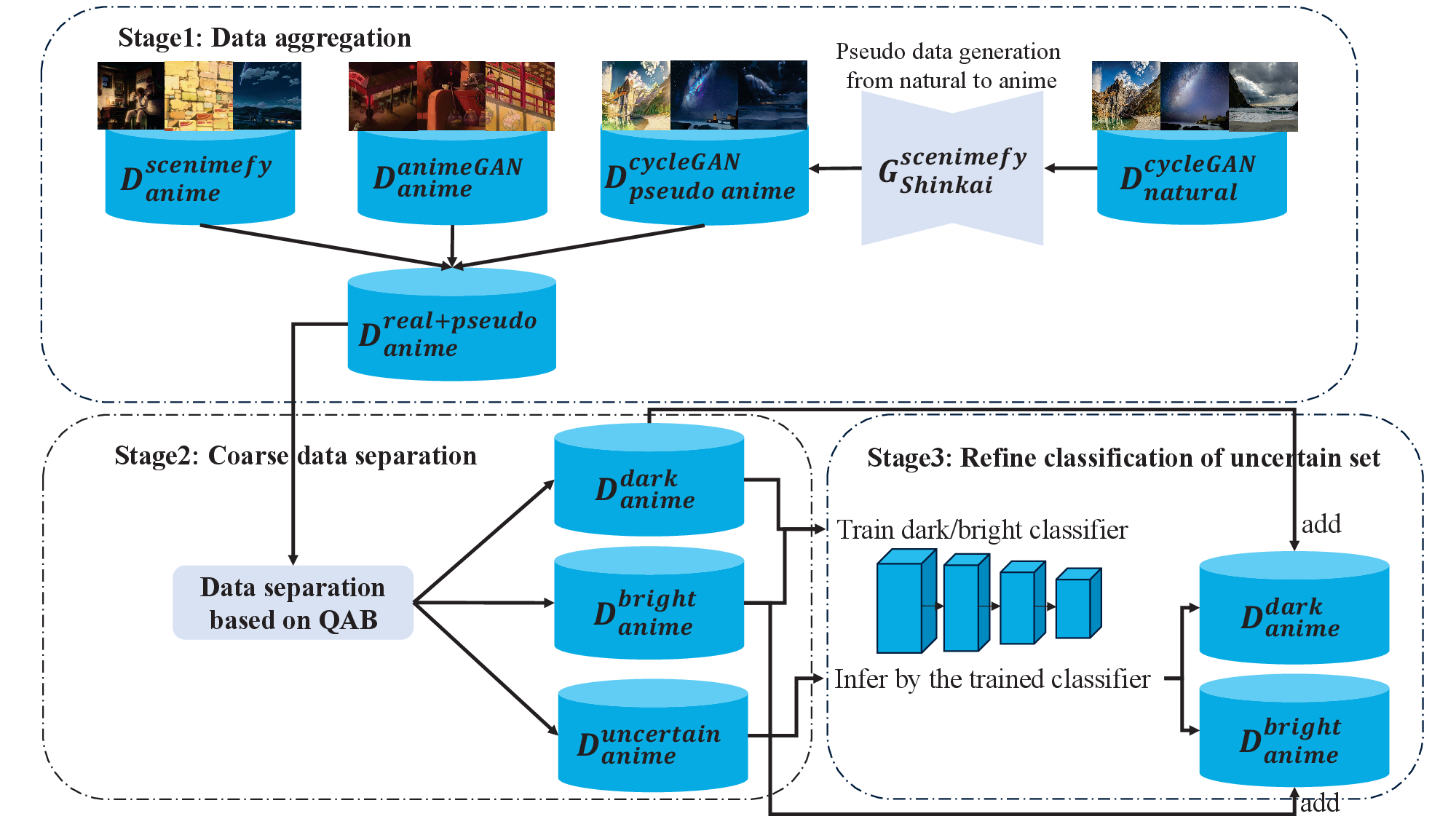}
	\caption{Overview of our data construction method for unpaired anime scenery dataset.}
	\label{Fig1}
\end{figure}

\section{Methodology}
\subsection{Unpaired Anime Data Construction Method}
\label{subsec:Unpaired Anime Data Construction Method}
To rescue the dilemma of data shortage, we devise a data construction method as illustrated in Fig.~\ref{Fig1}, consisting of three stages: data aggregation, coarse data separation and refine classification of uncertain set. 

In the first stage, for real anime scenery data, we gather 5952 and 6196 images respectively from Scenimefy~\cite{jiang2023scenimefy} and AnimeGAN~\cite{chen2020animegan}, denoted as $D_{anime}^{scenimefy}$ and $D_{anime}^{animeGAN}$. $D_{anime}^{scenimefy}$ is extracted from 9 Shinkai-style movies while $D_{anime}^{animeGAN}$ is extracted from 4 movies including Miyazaki, Shinkai and Hosoda styles. 

To further expand the data scale and diversity, we use a natural scenery set $D_{natural}^{cycleGAN}$ with 6656 images from CycleGAN~\cite{zhu2017unpaired} to generate their corresponding pseudo anime scenery set $D_{pseudo anime}^{cycleGAN}$ by employing the pre-trained natural-to-anime translation model $G_{Shinkai}^{scenimefy}$ of Scenimefy. 

Finally, the real anime data of $D_{anime}^{scenimefy}$ and $D_{anime}^{animeGAN}$ are aggregated along with the pseudo anime data of $D_{pseudo anime}^{cycleGAN}$ as a single set $D_{anime}^{real+pseudo}$ with totally 18804 images. 

To divide the collected anime data into dark or bright categories, the second stage initially filters the anime samples of $D_{anime}^{real+pseudo}$ into three subsets, i.e., $D_{anime}^{dark}$, $D_{anime}^{bright}$ and $D_{anime}^{uncertain}$. In contrast to $D_{anime}^{dark}$ and $D_{anime}^{bright}$ which contain confident dark and bright images, $D_{anime}^{uncertain}$ is composed of ambiguous images in the medium brightness range. We propound a quartile average brightness (QAB) algorithm to conduct data separation as follows:

\begin{equation}
	\small
	\left\{ \begin{array}{cl}
		I \in D_{anime}^{dark}& : \  mean(q(x,y))< B_{low},\exists q \in \left\{ I_{q1},...,I_{q4} \right\}\\
		I \in D_{anime}^{bright} & : \ mean(q(x,y))> B_{high},\forall q \in \left\{ I_{q1},...,I_{q4} \right\} \\
		I \in D_{anime}^{uncertain} & : otherwise\\
	\end{array} \right.
	\label{eq1}
\end{equation}

\noindent where $I$ is a candidate image loaded from $D_{anime}^{real+pseudo}$, while $D_{anime}^{dark}$, $D_{anime}^{bright}$ and $D_{anime}^{uncertain}$ are the resulting dark set, bright set and uncertain set separately. $mean(q(x,y))$ refers to the calculation of average brightness on the quartile patch $q(x,y)$ of candidate image $I$. $B_{low}$ and $B_{high}$ serve as the low and high boundary value for brightness. $\left\{ I_{q1},...,I_{q4} \right\}$ denotes the quartile patch set obtained by evenly splitting $I$. 

Concretely, we use quartile patch to measure the average brightness because the quartile can be the minimal unit that contains sufficient information to influence whether an image is dark or bright, similar to the quartile application in other studies~\cite{shih2016new,okizaki2023development}. 

We set $B_{low}$ and $B_{high}$ to 50 and 150 based on empirical observation of anime images and guided by the average brightness statistics reported in~\cite{shi2024maco}, which reliably separates dark and bright images and is robust to small variations. The average brightness calculation is written as below, where $w$ and $h$ denote the dimensions of the quartile patch. 

\begin{equation}
	mean(q(x,y))=\frac{1}{w\cdot h}\sum_{x=0}^{w-1}\sum_{y=0}^{h-1}q(x,y)
	\label{eq2}
\end{equation}

As the result of second stage, there exists 2958 dark images in $D_{anime}^{dark}$, 1848 bright images in $D_{anime}^{bright}$ and 13998 uncertain images in $D_{anime}^{uncertain}$. 

Furthermore, $D_{anime}^{dark}$ and $D_{anime}^{bright}$ can be taken for granted as the prior knowledge to distinguish between dark and bright classes, therefore, the third stage formulates a dark/bright classification task as: 
\begin{equation}
	\widehat{y}=F_{dark/bright}(x)=\omega_{dark/bright}\cdot x+\beta_{dark/bright}
	\label{eq3}
\end{equation}
where $\widehat{y}\in \left\{ dark,bright \right\}$ is the label, $x\in D_{anime}^{dark}\cup D_{anime}^{bright}$ is the dark/bright image. $\omega_{dark/bright}$ and $\beta_{dark/bright}$ are the weight and bias of classifier $F_{dark/bright}$. 

We adopt two sets of confident images, i.e., $D_{anime}^{dark}$ and $D_{anime}^{bright}$ to train the task which chooses ResNet18~\cite{he2016deep} for $F_{dark/bright}$ in our experiment. 

The trained classifier is used to further categorize the images of $D_{anime}^{uncertain}$ into $D_{anime}^{dark}$ or $D_{anime}^{bright}$, and a small amount of manual efforts is conducted to correct some of the medium-brightness samples with low confidence. 

After this process, $D_{anime}^{dark}$ and $D_{anime}^{bright}$ contain 10303 dark images and 8501 bright images, respectively. Following an 80/20 split, $D_{anime}^{dark}$ is divided into 8240 images for trainDark and 2063 images for testDark. $D_{anime}^{bright}$ with 8501 images is directly used as trainBright.

\subsection{Data Relativistic Uncertainty Framework}
\begin{figure}[h]
	\centering
	\includegraphics[width=\linewidth]{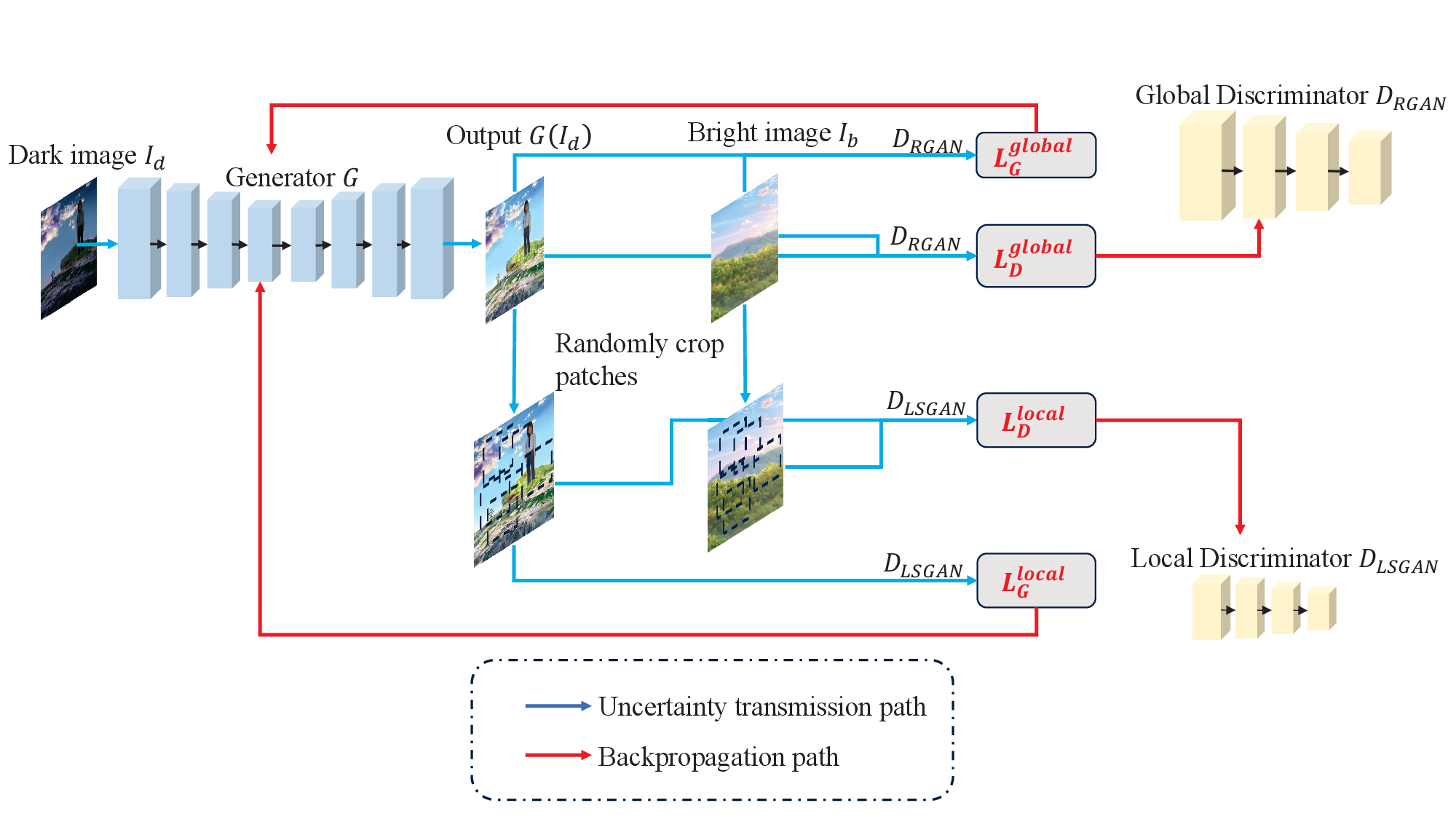}
	\caption{Visualize how illumination uncertainty of dark and bright images passes through each component in EnlightenGAN and finally affects the various loss terms of EnlightenGAN.}
	\label{Fig2}
\end{figure}
\vspace{+2.8em}
{\bf Motivation.} In practical application scenarios, illumination is always accompanied by uncertainty. Our training datasets including trainDark and trainBright are formed by most data from uncertain set $D_{anime}^{uncertain}$, which makes the data samples in trainDark and trainBright present some data uncertainty from the ideally dark and bright image. 

Conventionally, the training data for EnlightenGAN~\cite{jiang2021enlightengan} merely considers dark/bright images with high-confidence brightness~\cite{Chen2018Retinex,cai2018learning}, as is done in many LLIE methods~\cite{li2021low}. This practice limits the possibility of using data with comparatively uncertain brightness, which can be collected in a much easier way than with a cumbersome setup. 

If these methods attempt to learn directly with the dark/bright images with illumination uncertainty, they will overlook the illumination uncertainty and treat each sample with equal importance to update the loss terms. However, ignoring the illumination uncertainty from different samples will lead to unfair optimization from the updated loss terms and therefore, may cause non-smooth or under-exposed artifacts in moderately dark images (including dark images with partially bright areas) or extremely dark images after enhancement. 

The blue paths in Fig.~\ref{Fig2} exhibit how the illumination uncertainty information follows the dark ($I_{d}$) and bright ($I_{b}$) images through different networks and transformation modules, ultimately leading to the undesirable update of loss terms of EnlightenGAN. By tracing each blue path back to its origin, it is found that $L_{G}^{global}$, $L_{D}^{global}$ and $L_{D}^{local}$ are all affected by the illumination uncertainty of $I_{d}$ and $I_{b}$, while $L_{G}^{local}$ is only affected by the illumination uncertainty of $I_{d}$. 

To sum up, how to interpretably define the illumination uncertainty, quantify the illumination uncertainty of dark/bright images, and reform the loss terms for model are the keys to learning LLIE using illumination uncertainty data. 

{\bf Framework Design.} RaGAN has introduced a relativistic probability between the given real data and randomly sampled fake data for optimizing discriminator~\cite{jolicoeur2018relativistic}. Inspired by the relativistic probability concept of RaGAN, we propose a data relativistic uncertainty (DRU) framework to address the definition and quantification challenges of illumination uncertainty for dark/bright images, and reformulate the loss terms for EnlightenGAN to learn illumination uncertainty data. 

DRU framework defines the illumination uncertainty of a dark or bright image as the relativistic probability (RP) between the uncertain image and the ideally dark or bright image, which intuitively can be interpreted as the distance from the uncertain image to either the ideally dark or bright image. 

Although mathematically the distance between the uncertain image and the ideally dark or bright image can be calculated by existing metrics such as Euclidean distance and Cosine similarity, the ideally dark or bright image does not exist in reality and the results of these metrics cannot conform to the probability representation. Therefore, we quantify the RP between the uncertain image and the ideally dark/bright image using a learned probability network:
\begin{equation}
	RP = \left\{ \begin{array}{cl}
		RP_{d} & : F_{q}(x|y=0)=a_{q}(\omega_{q}\cdot x+\beta_{q})[0]\\
		RP_{b} & : F_{q}(x|y=1)=a_{q}(\omega_{q}\cdot x+\beta_{q})[1]
	\end{array} \right. 
\end{equation}
where $F_{q}$ denotes the probability network to quantify the RP of uncertain image $x$. $y=0$ means that $x$ is an uncertain dark image from trainDark with its RP marked as $RP_{d}$, while $y=1$ means that $x$ is an uncertain bright image from trainBright with its RP marked as $RP_{b}$. $F_{q}$ consists of the activation function $a_{q}$, weight $\omega_{q}$ and bias $\beta_{q}$. $[\cdot]$ denotes an array indexing operation based on $y$. In our practice, $F_{q}$ is implemented using a binary image classifier, as this is a straightforward approach to learn the genuine notion near ideal dark/bright image based on available prior knowledge from confident dark/bright images. We train the binary image classifier following the same way as the classifier $F_{dark/bright}$ in Sec.~\ref{subsec:Unpaired Anime Data Construction Method} and use a softmax function for $a_{q}$.

An alternative to implement $F_{q}$ can apply centriod-based methods~\cite{ren2024deep} to compute the two centroids of dark and bright clusters by using CNN feature maps extracted from confident dark and bright images, then quantify the probabilistic distance between the uncertain image and the dark/bright centroid (roughly replacing the ideal dark/bright image). However, this is not a learnable way based on prior knowledge, so it is unable to learn about the concept approaching ideal dark/bright image. 

To leverage the $RP_{d}$ and $RP_{b}$ quantified from uncertain dark and bright images to boost LLIE learning, we derive new illumination-aware loss terms for EnlightenGAN as follows:

\begin{equation}
	\small
	\hspace{-1em}
	L_{D}^{global}=RP_{b}\cdot RP_{d}[(D_{RGAN}(I_{b},G(I_{d}))-1)^{2}+D_{RGAN}(G(I_{d}),I_{b})^{2}]
\end{equation}

\begin{equation}
	\small
	L_{D}^{local}=RP_{b}\cdot (D_{LSGAN}(\overline{I_{b}})-1)^2+RP_{d}\cdot D_{LSGAN}(\overline{G(I_{d})})^2
\end{equation}

\begin{equation}
	\small
	\hspace{-1em}
	L_{G}^{global}=RP_{d}\cdot RP_{b}[(D_{RGAN}(G(I_{d}),I_{b})-1)^2+D_{RGAN}(I_{b},G(I_{d}))^2]
\end{equation}

\begin{equation}
	\small
	L_{G}^{local}=RP_{d}\cdot (D_{LSGAN}(\overline{G(I_{d})})-1)^2
\end{equation}
where $L_{D}^{global}$ and $L_{D}^{local}$ are respectively the global loss function for discriminator $D_{RGAN}$~\cite{jolicoeur2018relativistic} and the local loss function  for discriminator $D_{LSGAN}$~\cite{mao2017least}. $L_{G}^{global}$ and $L_{G}^{local}$ are the global and local loss functions for generator $G$.
As can be seen in Fig.~\ref{Fig2}, $I_{b}$ denotes an uncertain bright image from trainBright while $I_{d}$ is an uncertain dark image from trainDark. Accordingly, $G(I_{d})$ is the enhanced image of $I_{d}$ after passing through $G$. $\overline{I_{b}}$ and $\overline{G(I_{d})}$ refer to a randomly cropped patch from $I_{b}$ and $G(I_{d})$. Here we quantify $RP_{d}$ and $RP_{b}$ for each image of trainDark and trainBright, since even for high-confidence images, there is always tiny illumination uncertainty from the ideal dark and bright image. 

Different from the primitive loss terms of EnlightenGAN~\cite{jiang2021enlightengan}, the nature of our illumination-aware loss terms tends to allocate stronger gradient optimization towards dark/bright samples with lower illumination uncertainty, and conversely, weaker gradient optimization towards dark/bright samples with higher illumination uncertainty. This allows to exploit the illumination uncertainty as prior information to adaptively optimize model training based on the various illumination conditions of dark-bright image pairs. Additionally, Self Feature Preserving (SFP) losses~\cite{jiang2021enlightengan} are not affected by the illumination uncertainty of dark and bright images. Hence, we remain them to constrain the perceptual similarity between $I_{d}$ and $G(I_{d})$. Finally, EnlightenGAN aims to minimize the entire training objective:

\begin{equation}
	\small
	L_{overall}=\overset{Losses\ from\ DRU}{\overbrace{L_{D}^{global}+L_{D}^{local}+L_{G}^{global}+L_{G}^{local}}}+L_{SFP}^{global}+L_{SFP}^{local}
\end{equation}

{\bf Interpretability of DRU.} DRU can be understood by analogy to the wave-particle duality of light~\cite{milonni1984wave}. Each sample is transformed into a representation similar to a wave function via a probability quantizer, with two possible states: bright and dark. The corresponding $RP$ indicates the probability of each state.

When the loss function treats a sample as a specific state, this is analogous to observing a definite particle state. Since the probability reflects the illumination level, the loss for each state can be weighted by its corresponding probability, thus calibrating each sample’s contribution to the loss update. Without DRU, the loss function assumes a 100\% probability for a specific state, ignoring the actual illumination uncertainty. This leads to illumination bias propagating through the loss updates, causing under- or over-enhancement in the resulting model.

\section{Experiments}
\subsection{Implementation Details}
Regarding the DRU framework, we implement the probability network $F_{q}$ by six choices of modern backbone networks including ResNet18, ResNet50~\cite{he2016deep}, MobileNetV2~\cite{sandler2018mobilenetv2}, VGG13BN, VGG19~\cite{simonyan2014very} and ViT-B16~\cite{dosovitskiy2020image}. These backbone networks follow the identical training data and recipe of the classifier $F_{dark/bright}$ as described in Sec.~\ref{subsec:Unpaired Anime Data Construction Method}. The training recipe uses the pre-trained weights from ImageNet~\cite{deng2009imagenet}, a batch size of 64 to run over 80 epochs, and the optimizer SGD~\cite{robbins1951stochastic} with learning rate of $1e^{-3}$ and momentum of 0.9. 

After obtaining the trained backbone networks to quantify the illumination uncertainty, we employ EnlightenGAN as the base architecture while retaining its default training hyperparameters~\cite{jiang2021enlightengan}, and train six versions of EnlightenGAN using the DRU framework. We denote each version as DRU-EnlightenGAN-Backbone in the experiments. Our models are compared with five state-of-the-art (SOTA) unsupervised LLIE methods, such as Vanilla EnlightenGAN~\cite{jiang2021enlightengan}, SCI~\cite{ma2022toward}, ZeroDCE++~\cite{li2021learning}, RUAS~\cite{liu2021retinex} and CLIP-LIT~\cite{liang2023iterative}. 

To avoid the domain gap problem and ensure fair comparisons, trainDark and trainBright of the unpaired anime scenery dataset are adopted for all the experimental trainings. Specifically, our models, Vanilla EnlightenGAN and CLIP-LIT use trainDark and trainBright as unpaired dark-bright training sets, whereas, ZeroDCE++ combines trainDark and trainBright into a multi-exposure training set. SCI and RUAS use trainDark but not trainBright since they only need low-light training images. The training settings of SOTA methods follow the original practices of~\cite{jiang2021enlightengan,ma2022toward,li2021learning,liu2021retinex,liang2023iterative}. Lastly, testDark in the unpaired anime scenery dataset is utilized to evaluate the enhancement performance of trained models and the test results are concretely measured by four indicators involving BRISQUE~\cite{mittal2012no}, PIQE~\cite{venkatanath2015blind}, PI~\cite{blau20182018} and NIMA~\cite{talebi2018nima}. BRISQUE and PIQE assess the perceptual quality relative to the distortion level. PI measures the visual quality along with the reconstruction accuracy. NIMA estimates the aesthetic quality. Additionally, we conduct a user study to evaluate the perceptual and aesthetic quality of different models.
\subsection{State-of-the-Art Comparison}

\begin{table}[]
	\setlength{\tabcolsep}{1.2mm}
	\caption{Averaged BRISQUE, PIQE, PI and NIMA scores on the test set. ↓ indicates smaller score is better while ↑ indicates higher score is better. Input denotes the test images. Bold means the best score and underline means the second. All the results of SOTA method are implemented using the official codes.}
	\scalebox{0.9}{%
		\begin{tabular}{cllll}
			\hline
			Method                       & \multicolumn{1}{c}{BRISQUE↓} & \multicolumn{1}{c}{PIQE↓} & \multicolumn{1}{c}{PI↓} & \multicolumn{1}{c}{NIMA↑} \\ \hline
			Input                        & 32.4483                      & 48.7356                   & 4.9890                   & 4.7587                    \\
			SCI                          & 27.8680                       & 46.0016                   & 4.4208                  & 4.6787                    \\
			ZeroDCE++                    & 28.4048                      & 46.1118                   & 4.4507                  & 4.4888                    \\
			RUAS                         & 49.1256                      & 64.5322                   & 6.6119                  & 4.1766                    \\
			CLIP-LIT                     & 29.2622                      & 44.3198                   & 4.4435                  & 4.6900                      \\
			Vanilla EnlightenGAN         & 27.2812                      & 45.1133                   & 4.3932                  & 4.7456                    \\
			DRU-EnlightenGAN-ResNet18    & {\ul 25.9728}                & \textbf{41.9577}          & 4.3208                  & 4.7602                    \\
			DRU-EnlightenGAN-ResNet50    & \textbf{25.9235}             & 42.7084                   & {\ul 4.3175}            & 4.7735                    \\
			DRU-EnlightenGAN-MobileNetV2 & 26.5647                      & {\ul 42.6536}             & 4.3207                  & 4.7726                    \\
			DRU-EnlightenGAN-VGG13BN     & 26.3275                      & 43.7998                   & 4.3497                  & {\ul 4.7776}              \\
			DRU-EnlightenGAN-VGG19       & 26.5520                       & 43.5011                   & 4.3408                  & 4.7576                    \\
			DRU-EnlightenGAN-ViT-B16     & 26.4526                      & 42.7171                   & \textbf{4.3032}         & \textbf{4.7901}           \\ \hline
	\end{tabular}}
	\label{tab1}
\end{table}

\begin{figure}[h]
	\centering
	\includegraphics[width=\linewidth]{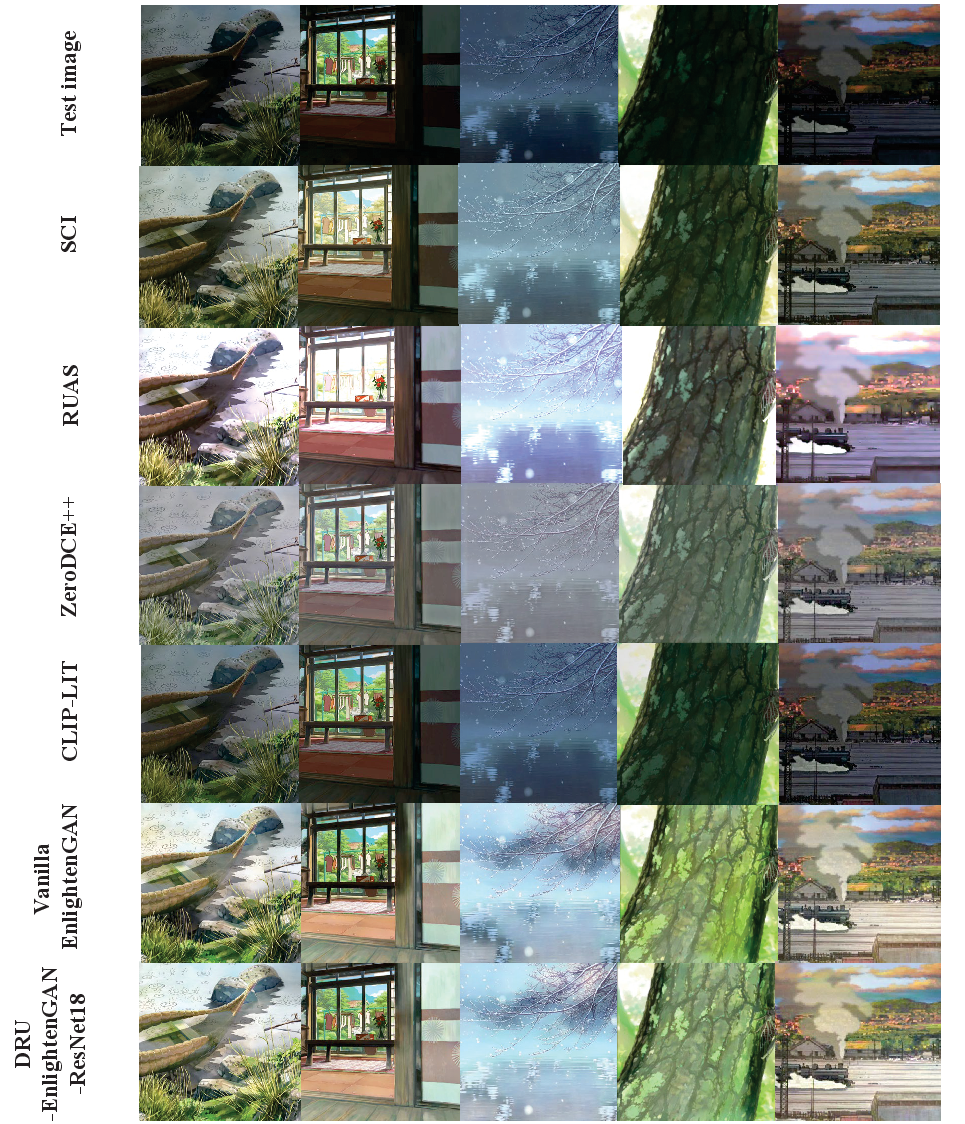}
	\caption{Qualitative comparisons among different methods. Zoom in for detailed view.}
	\label{Fig3}
\end{figure}

\textbf{Qualitative analysis.} We compare the qualitative results of DRU-EnlightenGAN-ResNet18 with other SOTA methods as shown in Fig.~\ref{Fig3}. SCI, ZeroDCE++ and CLIP-LIT present inadequate results with varying degrees of dark coverage. ZeroDCE++ looks slightly better than the other two, but exists a gray color bias. Although the results of RUAS are brighter than all other methods, they suffer from overexposed artifacts, leading to a noticeable blue color cast. Last two rows exhibit more pleasing visual results than their counterparts and furthermost, DRU-EnlightenGAN-ResNet18 outperforms Vanilla EnlightenGAN with less shadow and color bias. Here, Fig.~\ref{Fig3}(row,column) is used to denote the image position. 

The local region between boat and grass in Fig.~\ref{Fig3}(6,1) displays darker and larger shadow than that of Fig.~\ref{Fig3}(7,1). The similar phenomena occur on the pillar in Fig.~\ref{Fig3}(6,2) and Fig.~\ref{Fig3}(7,2), the left part of lake in Fig.~\ref{Fig3}(6,3) and Fig.~\ref{Fig3}(7,3), and the right region of sky in Fig.~\ref{Fig3}(6,5) and Fig.~\ref{Fig3}(7,5). Besides that, the tree of Fig.~\ref{Fig3}(6,4) shows a yellow cast while that of Fig.~\ref{Fig3}(7,4) is naturally close to green. Both of them preserve clearer foreground textures and more complete background details compared with other methods. Therefore, the above visual analysis sufficiently proves the effectiveness of DRU framework. 

\textbf{Quantitative analysis.} By referring to Table~\ref{tab1}, the DRU versions of EnlightenGAN basically surpass all the SOTA counterparts across four indicators by a large margin, which further demonstrates the superiority and practicality of DRU framework. The aim of low-illumination enhancement for anime scenery images is not simply to improve or raise the brightness. Even though RUAS attains the brightest results in Fig.~\ref{Fig3}, it performs the bottom in all metrics, which reveals that RUAS is unable to learn a rational illumination concept from the low-light images in trainDark, aligning with the discovery in~\cite{wang2024zero}. Note that Vanilla EnlightenGAN achieves better scores than other SOTA methods, followed by SCI. 

For simplicity, we refer to each DRU version of EnlightenGAN by its backbone below. ResNet50 and ResNet18 shows the first and second score for BRISQUE. ResNet18 and MobileNetV2 are ranked as the first and second for PIQE. ViT-B16 is the only one that reaches the best scores for PI and NIMA, respectively followed by ResNet50 and VGG13BN. 

Consequently, DRU-EnlightenGAN-ViT-B16 stands out among all the methods in terms of overall quality scores. In particular, the self-attention mechanism in Transformer enables direct modeling of relationships between any positions in the image, allowing it to capture both local and non-local illumination variations. Compared with CNN-based backbones that rely primarily on local receptive fields, Transformer effectively models global luminance distribution. This capability is especially beneficial for low-illumination anime scenery images, where brightness variations can arise at multiple scales, from local dark patches to large areas of uneven illumination. Therefore, the global representation capability of ViT-B16 contributes to its superior performance on PI and NIMA metrics, implying that the Transformer structure can exert a stronger generalization capacity than CNN in quantifying illumination uncertainty. 

\begin{table}[t]
	\centering
	\caption{The user study results are reported as average preference scores (APS), representing the proportion of times each method is selected as the best. Bold indicates the best score, and underline indicates the second best.}
	\label{tab:user_study}
	\begin{tabular}{c c}
		\hline
		Method & \makebox[2cm][r]{APS (\%)} \\
		\hline
		SCI & \makebox[2cm][r]{6.5650} \\[1pt]
		RUAS & \makebox[2cm][r]{14.3770} \\[1pt]
		ZeroDCE++ & \makebox[2cm][r]{6.5650} \\[1pt]
		CLIP-LIT & \makebox[2cm][r]{5.0030} \\[1pt]
		Vanilla EnlightenGAN & \makebox[2cm][r]{\underline{33.1270}} \\[1pt]
		DRU-EnlightenGAN-ResNet18 &\makebox[2cm][r]{ \textbf{34.3790}} \\
		\hline
	\end{tabular}
\end{table}

{\textbf{User study.}} We conduct a user study with 32 participants to evaluate the visual quality of anime scenes produced by each method. Participants select their preferred options from six candidates across 10 image sets, according to three criteria: appropriate brightness, naturalness of colors, and adherence to anime aesthetics. As summarized in Table~\ref{tab:user_study}, the average preference scores show that DRU-EnlightenGAN-ResNet18 outperforms all other methods, demonstrating the effectiveness of our framework in refining EnlightenGAN.

\subsection{Ablation Study}
In this section, two initial ablation studies are conducted using DRU-EnlightenGAN-ResNet18 and Vanilla EnlightenGAN, hereafter referred to as DRU and Vanilla, to investigate two key questions. Details of the remaining experiments are provided in the following sections.

\textbf{Are both confident and uncertain dark/bright images necessary for DRU training?} To examine the effect of confident and uncertain images to DRU and Vanilla training respectively, we separate the confident and uncertain images from trainDark and trainBright, which are denoted as Confident and Uncertain for training. So Confident+Uncertain is equivalent to all the images of trainDark and trainBright. The goal of enhancement task takes into account both perceptual and aesthetic qualities, namely, the results for BRISQUE and NIMA are expected to approach lower and higher respectively. From this view, DRU on Confident+Uncertain demonstrates the best trade-off between the perceptual and aesthetic qualities in Table~\ref{tab2}. 

With only Confident, DRU performs well in BRISQUE but poorly in NIMA, indicating that the enhanced results are bright but not smooth and aesthetically pleasing. However, using uncertainty only for DRU leads to relatively dark but smooth enhanced results. Thus, both confident and uncertain images are needed for DRU training since confident images are strong priors for DRU to learn towards better perceptual quality while uncertain images benefit DRU from the illumination uncertainty to learn greater aesthetic quality. 

Although the overall performance of Vanilla is suboptimal to DRU. Interestingly, we observe from Table~\ref{tab2} that Vanilla obtains better BRISQUE and NIMA scores on Uncertain than another two settings, followed by Confident+Uncertain. This indicates the importance of uncertain images for Vanilla training and the necessity of DRU to exploit illumination uncertainty of confident and uncertain images for further learning.

\begin{table}[]
	\caption{Performance of DRU and Vanilla trained on different data settings. Confident or Uncertain means only using the confident or uncertain images of trainDark and trainBright. Confident+Uncertain is to use all the images of trainDark and trainBright. We report the scores on testDark as BRISQUE↓$|$NIMA↑, namely the left is BRISQUE(lower is better) and the right is NIMA(higher is better).}
	\begin{tabular}{ccc}
		\hline
		Data setting & DRU & Vanilla \\ \hline
		Confident & 24.7172$|$4.5146 &  25.5591$|$4.5620\\
		Uncertain & 26.7868$|$4.7715 & 27.2191$|$4.8069\\
		Confident + Uncertain  & 25.9728$|$4.7602 & 27.2812$|$4.7456 \\ \hline
	\end{tabular}
	\label{tab2}
\end{table}

\textbf{Can DRU mitigate the impact of training data noise caused by misclassification?}
\begin{table}[]
	\caption{We inspect the ability of DRU to mitigate the harmful effect of data noise caused by misclassification. Original denotes the reliable trainDark and trainBright in Sec.~\ref{subsec:Unpaired Anime Data Construction Method}. Noisy denotes the incorrect trainDark and trainBright that are reclassified using a ResNet18 pretrained on ImageNet. We report the scores on testDark as BRISQUE↓$|$NIMA↑, i.e., the left is BRISQUE(lower is better) and the right is NIMA(higher is better).}
	\centering
	\begin{tabular}{ccc}
		\hline
		Data set & DRU & Vanilla \\ \hline
		Original & 25.9728$|$4.7602 & 27.2812$|$4.7456 \\
		Noisy & 26.8293$|$4.7996 & 29.1476$|$4.8016 \\ \hline
	\end{tabular}
	\label{tab3}
\end{table}
We reclassify the original trainDark and trainBright using an unreliable pretrained ResNet18 to produce new trainDark and trainBright with severe data noise. Table~\ref{tab3} shows the performance of DRU and Vanilla after training on Original and Noisy sets. 

Between the Original and Noisy sets, the BRISQUE difference of DRU is 0.8565, which is much smaller than the 1.8664 of Vanilla. The NIMA difference of DRU is 0.0394, less than the 0.056 of Vanilla. Vanilla trained on Noisy produces darker enhanced results that are given the higher BRISQUE score, and darker images tend to exhibit more smooth textures, affecting its NIMA score higher. So this signifies worse performance. DRU achieves better trade-off between BRISQUE and NIMA scores and less variation for each metric. Hence, DRU is able to reduce the performance loss raised by data noise due to misclassification compared to Vanilla. 

\begin{table}[htbp]
	\centering
	\caption{Performance of different methods on the unseen dataset. Input denotes the test images. We report the scores as BRISQUE↓$|$NIMA↑. Vanilla and DRU-ViT-B16 correspond to two different versions of EnlightenGAN. Bold marks the best score, underline marks the second best.}
	\label{tab:unseen_eval}
	\renewcommand{\arraystretch}{1.2} 
	\begin{tabular}{c c} 
		\hline 
		Method & \hspace{-1.1em}BRISQUE↓$|$NIMA↑ \\
		\hline 
		Input & 37.8164
		  $|$ 3.9054 \\
		SCI & \underline{31.6215}
		  $|$ 3.8296
		   \\
		RUAS & 35.4778
		  $|$ 3.5323
		   \\
		ZeroDCE++ & 33.9296
		  $|$ 3.6256
		   \\
		CLIP-LIT & 33.1846
		  $|$ 3.8368
		   \\
		Vanilla & 32.0589
		  $|$ \underline{3.9199}
		   \\
		DRU-ViT-B16 & \textbf{31.1713
		  $|$ 3.9299}
		   \\
		\hline 
	\end{tabular}
\end{table}

{\textbf{Evaluation on unseen dataset.}} To further evaluate the generalizability of different methods, we construct an unseen dark dataset of 738 anime scenery images sourced from~\cite{zhu2025animedl,deepghs_danbooru2024_sfw,kon_paprika2006}. As shown in Table~\ref{tab:unseen_eval}, DRU-ViT-B16 obtains the best performance among all methods, indicating robust perceptual and aesthetic generalization on unseen data.

\begin{table}[htbp]
	\centering
	\caption{Performance of RUAS variants on testDark and unseen datasets. Scores are reported as BRISQUE↓$|$NIMA↑. RUAS is the vanilla variant. RUAS-DRU denotes the RUAS variant trained with DRU-ResNet18. Bold indicates the best score.}
	\label{tab:ruas_variants}
	\renewcommand{\arraystretch}{1.2} 
	\begin{tabular}{c c c c c} 
		\hline
		Variant & \multicolumn{1}{c}{testDark} & \multicolumn{1}{c}{unseen} \\
		\hline
		RUAS       & 49.1256
		$|$ 4.1766  & 35.4778
		$|$ 3.5323 \\
		RUAS-DRU   & \textbf{34.6225
		$|$ 4.2826}
		  & \textbf{35.4175
		$|$ 3.5390 }
		 \\
		\hline
	\end{tabular}
	
\end{table}

{\textbf{Transferability of DRU.}} To assess the generality of our framework, we adapt DRU to train a variant of RUAS. Table~\ref{tab:ruas_variants} compares the performance of RUAS and RUAS-DRU on testDark and the unseen dataset. Training RUAS with DRU yields notable improvements in both perceptual and aesthetic quality, highlighting the framework’s effectiveness and its potential applicability to low-light enhancement beyond EnlightenGAN.

\begin{figure}[h]
	\centering
	\includegraphics[width=\linewidth]{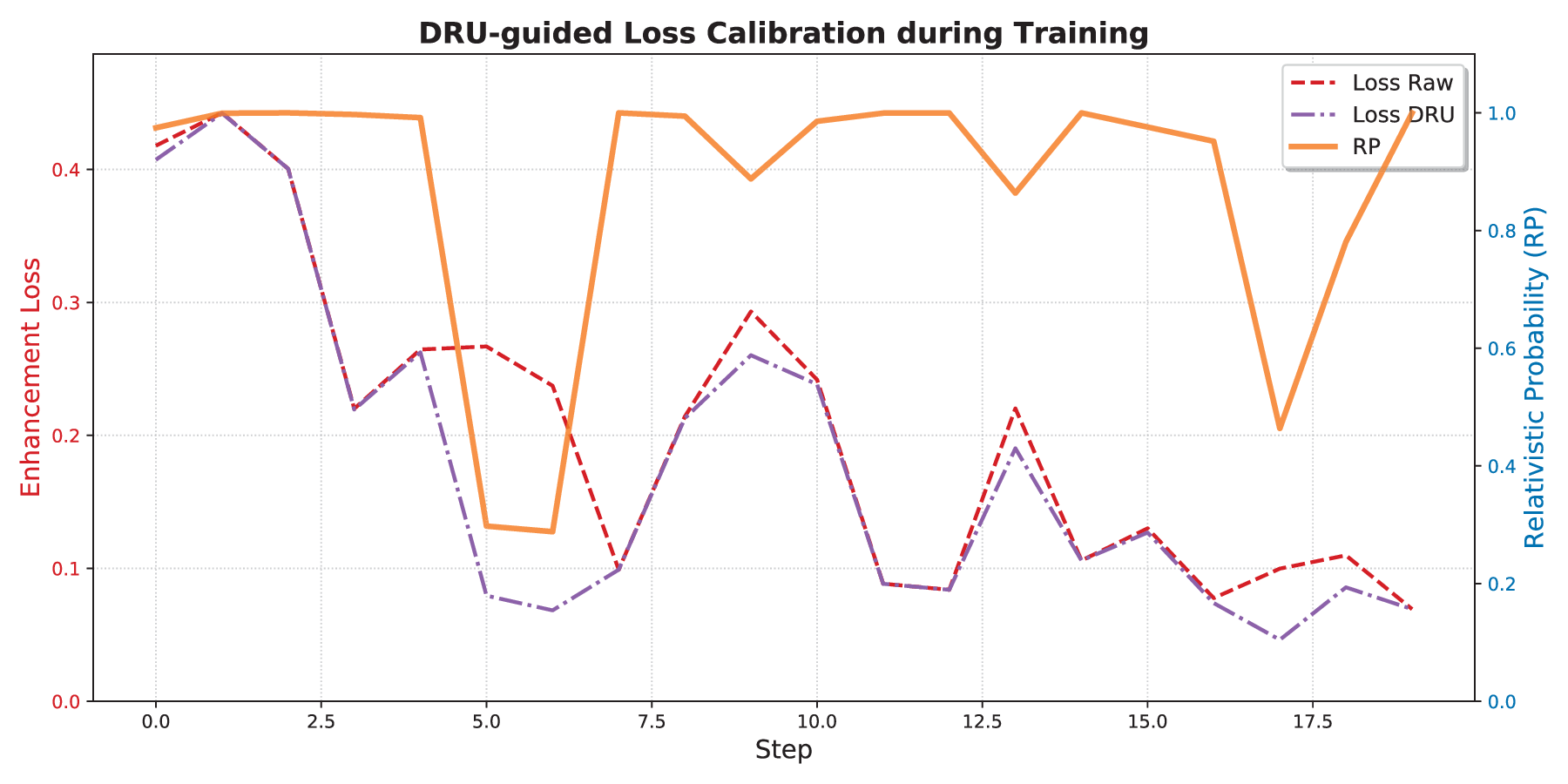}
	\caption{Training dynamics of the baseline loss (Loss Raw, red) and the DRU-weighted loss (Loss DRU, purple), together with the relativistic probability (RP, orange). Each training step corresponds to a single image due to a batch size of 1. For confident images with high RP values, the DRU-weighted loss closely follows the baseline loss. In contrast, for uncertain images with low RP values, the DRU mechanism suppresses the loss magnitude, thereby limiting their contribution to optimization.}
	\label{Fig5}
\end{figure}

\textbf{Visualization of DRU mechanism.} To analyze the behavior of the DRU framework, we train RUAS on a mixture of confident and uncertain samples. Fig.~\ref{Fig5} illustrates the evolution of the relativistic probability (RP), baseline loss, and DRU-weighted loss over training. With a batch size of 1, each step corresponds to a single image. For confident samples (high RP), the DRU-weighted loss closely tracks the baseline loss, indicating that these samples are more trusted during optimization. Conversely, for uncertain samples (low RP), the DRU mechanism attenuates the loss, reducing their contribution and preventing unreliable samples from disproportionately influencing training. In this way, the DRU mechanism enables the model to effectively learn from samples with varying levels of illumination uncertainty.

\section{Conclusion}
We address the domain gap and data shortage problems faced by low-illumination enhancement for anime scenery images through establishing the first unpaired anime scenery dataset. In addition, we propose a Data Relativistic Uncertainty (DRU) framework that reformulates low-light enhancement learning from a data uncertainty perspective, enabling training with uncertain samples rather than relying solely on confident ones. Our framework is applicable not only to EnlightenGAN, but also to any state-of-the-art methods as long as their loss terms can be adapted to use the framework. Experimental results evince its advantageous performance across multiple quality metrics. Our framework provides the possibility for the realization of future work in~\cite{jiang2021enlightengan}. We also expect that this new paradigm of data uncertainty learning can generalize to other vision and language tasks.


\section*{Acknowledgement}
We would like to thank all participants of the user study for their interest and support in this research. We also extend our gratitude to the first author of Scenimefy for her help in processing and sharing her data, as well as to other authors who provided additional relevant data.


{\small
\bibliographystyle{cvm}
\bibliography{cvmbib}
}

\end{document}